\pdfoutput=1
\documentclass[11pt]{article}
\usepackage[preprint]{acl}
\usepackage{times}
\usepackage{pdfpages}
\usepackage{latexsym}
\usepackage[T1]{fontenc}
\usepackage[utf8]{inputenc}
\usepackage{CJKutf8}
\usepackage{csquotes}
\usepackage{microtype}
\usepackage{inconsolata}
\usepackage{adjustbox}
\usepackage{makecell}
\usepackage{graphicx} 
\usepackage{lscape}
\usepackage{multirow}
\usepackage{amssymb}
\usepackage{listings}
\usepackage{pdfpages}
\usepackage{tikz}
\usepackage{colortbl}
\usepackage{booktabs}
\usepackage{amsmath}
\usepackage{xspace}
\usepackage{booktabs}     
\usepackage{tabularx} 
\usepackage{float}
\usepackage{CJKutf8}
\usepackage{ulem}
\usepackage{array}
\usepackage{fontawesome}
\newcolumntype{L}[1]{>{\raggedright\arraybackslash}m{#1}}
\newcolumntype{C}{>{\centering\arraybackslash}c}
\newcolumntype{M}[1]{>{\centering\arraybackslash}m{#1}}

\newcommand{\stkout}[1]{\ifmmode\text{\sout{\ensuremath{#1}}}\else\sout{#1}\fi}
\definecolor{lightblue}{RGB}{173,216,230}
\title{Moderation Matters: \\ Measuring Conversational Moderation Impact in English as a Second Language Group Discussion}  

\author{Rena Gao, Ming-Bin Chen, Lea Frermann, Jey Han Lau \\  
The University of Melbourne, Parkville, 3052, Australia \\ \texttt{\{rena.gao,mingbin,lea.frermann\}@unimelb.edu.au}, \texttt{jeyhan.lau@gmail.com}
}

\begin{document}
\maketitle

\begin{abstract}
English as a Second Language (ESL) speakers often struggle to engage in group discussions due to language barriers. While moderators can facilitate participation, few studies assess conversational engagement and evaluate moderation effectiveness. To address this gap, we develop a dataset comprising 17 sessions from an online ESL conversation club, which includes both moderated and non-moderated discussions. We then introduce an approach that integrates automatic ESL dialogue assessment and a framework that categorizes moderation strategies. Our findings indicate that moderators help improve the flow of topics and start/end a conversation. Interestingly, we find active acknowledgement and encouragement to be the most effective moderation strategy, while excessive information and opinion sharing by moderators has a negative impact. Ultimately, our study paves the way for analyzing ESL group discussions and the role of moderators in non-native conversation settings. Code and data are available at \url{https://github.com/RenaGao/L2Moderator}.

%from 17 sessions involving 42 Asian ESL speakers ~\citep{jones1999silence} ~\citep{kayi2013scaffolding}
\end{abstract}

\begin{figure}[t]
    \centering
    \includegraphics[width=\linewidth]{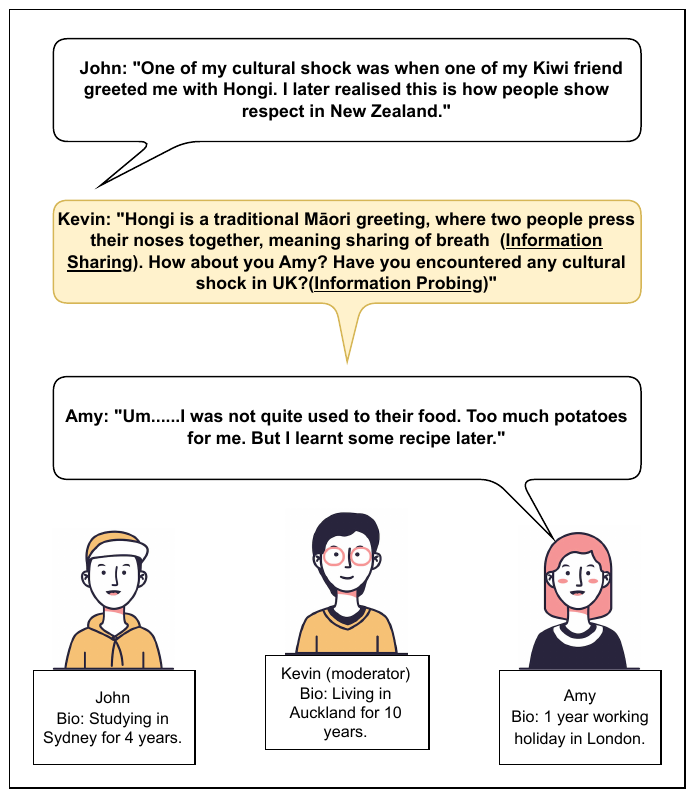}
    \caption{Example of a moderated conversation involving ESL Asian speakers. Highlighted dialogue bubbles represent moderator interventions, while underlined text indicates the tags of the moderation strategies.}
    \label{fig:slomo_demo}
\end{figure}

\section{Introduction}\label{sec:intro} 

Participation in group discussions has been widely recognized as an effective means for language acquisition ~\citep{crisianita2022use, pica1985role, hudgins1986teaching}. However, numerous studies across diverse fields have highlighted the challenges encountered by English as a Second Language (ESL) learners—particularly those from Asian linguistic and cultural backgrounds—when interacting in English group discussions~\citep{king2013silence, lee2009speaking, li2006don, yang2010doing}. As illustrated in Figure \ref{fig:slomo_demo}, introducing a moderator into the discussion has been identified as an effective solution to enhance ESL speakers’ participation ~\citep{hamzah2016effect}. Moderators assist through various interventions such as providing guidance~\citep{hamzah2016effect}, which help to support engagement~\citep{reddington2018managing} and communication~\citep{vasodavan2020moderation}. Existing ESL assessment tools predominantly focus on evaluating writing skills and lack in support for dynamic spoken interactions, especially in group settings. While recent studies have begun exploring the measurement of interaction and engagement in two-party second language conversations~\citep{gao-etal-2025-interaction, gao2024cnima}, a notable gap remains: how to assess engagement and interaction in multi-party ESL discussions. Furthermore, it's unclear how moderator interventions influence ESL group dynamics.

In this study, we seek to quantify the impact of moderators on participants' dialogue quality directly from dialogue transcripts (without relying on external measures such as questionnaires). Specifically, we investigate whether moderator presence influences performance and which moderation strategies are most effective to stimulate discussion. To this end, we develop a dataset of 17 ESL group discussions, covering both moderated and non-moderated conditions. Building on recent advancements in ESL dialogue evaluation~\citep{gao2024cnima} and conversational moderation analysis~\citep{chen2024whow}, we develop an approach
that automates the assessment of ESL group discussions based on the transcripts and examines moderation strategies by categorizing them into ten distinct types tailored for ESL contexts.  We find that %moderation interventions associated with social motivation, especially “Echoing”, were the most effective in ESL group discussions, statistically demonstrating that for non-native ESL speakers,
active acknowledgment and encouragement from the moderator is the most effective strategy to improve ESL discussions, while excessive information or opinion sharing has the opposite effect.
%and guide future educational NLP on dialogue moderator design for language online education.
In sum, our contributions are:

 \begin{itemize}
     \item We introduce the first moderated multi-party ESL group discussion conversation dataset, comprising 17 sessions (total 16.5 hours of recorded transcripts and 9,843 sentences), which includes parallel moderated and non-moderated sessions on the same set of topics.
     \item By integrating the dialogue quality assessment approach from \citet{gao2024cnima} with the WHoW moderation analysis framework from \citet{chen2024whow}, we introduce a novel method that offers an automatic, fine-grained evaluation of dialogue quality and moderator intervention in ESL group discussions.
     \item Comparative analysis reveals that the presence of moderator significantly improves topic management and the quality of conversation openings and closings. The most effective moderator intervention is active acknowledgement and encouragement; and the least is information/opinion sharing.
     %the “Echoing” intervention  emerged as the most effective, while “Opinion Sharing” the least.
 \end{itemize}

\section{Related work}\label{sec:related-work}

\paragraph{Barriers to participation in group discussions for ESL speakers}  

% English-Second-Language (ESL) speakers often encounter significant communication and interaction barriers primarily due to cross-language ability challenges~\cite{higuchi2023impact}. Unlike native speakers, who often have a more intuitive grasp of language nuances, ESL speakers may struggle with constructing sentences that accurately convey their intended meaning \cite{turnbull2015effects}, convert accurate ideas~\cite{hessel2017new}, struggle to start the talk in a second language~\cite{tan2020influence}, and thus with starting and contributing to group conversations~\cite{sampson2024emergence}. This can lead to misunderstandings in interaction and hinder the conversation flow~\cite{lam2006gauging}, as ESL speakers might need extra time to formulate their utterances facing spoken interactions. It can complicate interactions, particularly in informal or spontaneous conversations.

ESL speakers often encounter significant barriers to effective communication and participation in group discussions, stemming from cross-language ability challenges~\cite{higuchi2023impact}. Unlike native speakers, ESL speakers may struggle to construct sentences that accurately convey their intended meaning and ideas~\cite{turnbull2015effects, hessel2017new}. They often require additional time to formulate responses in spoken interactions~\cite{lam2006gauging}, which can be particularly problematic in informal discussions where quick exchanges are expected. Moreover, ESL speakers are less likely to initiate conversations~\cite{tan2020influence}, further limiting their participation. Above barriers disrupt the flow of conversations, increase the risk of misunderstandings, and significantly reduce ESL speakers' engagement in group discussions~\cite{sampson2024emergence,gao2024listenership}.

%Additional factors, such as cultural differences, low self-confidence, social anxiety, and limited motivation to communicate~\citep{jones1999silence, de2009learners, norton2014non, , mak2011exploration}, exacerbate this reluctance. 

Effectively participating in group interactions is essential for ESL speakers~\cite{rao2019enhancing}. Interactive conversations provide ESL speakers with exposure to the practical use of language, allowing them to learn how to naturally communicate~\cite{gao2024listenership,rao2019enhancing}. This learning cannot be fully replaced by passive methods such as reading or listening~\cite{wu2025data,maarof2018effect}. Consequently, ESL speakers face a dilemma: improving their language skills requires active participation, yet their limited proficiency and confidence often undermines their ability to engage effectively. To address this, educational strategies such as structured group discussions and explicit guidance are commonly employed to facilitate meaningful participation~\cite{pino2010english,webster2012teaching}.

% Effectively practising those interaction skills is crucial for ESL speakers to overcome these barriers~\cite{rao2019enhancing}. Interaction exposes ESL speakers to the practical use of language in various contexts, enabling them to learn and internalize how native speakers naturally communicate \cite{gao2024listenership,rao2019enhancing}. Without active engagement, ESL speakers may find it difficult to reach the level of fluency necessary for effective communication~\cite{maarof2018effect}
% , as passive learning methods like reading or listening alone do not provide the same opportunities to apply language skills in real-time situations~\cite{dai2023second,wu2025data}. During second language learning, these strategies need further instructed from language educators, especially under group discussions to be equipped with explicit guidance~\cite{pino2010english,webster2012teaching}.  Therefore, effective moderation among ESL speakers' interaction can prominently improve their ability in second language communication, language learning, and language interactional competence.  

\paragraph{Dialogue quality evaluation}

% Evaluating the quality of conversational dialogue and justifying the evaluation in an explainable way has been a long-standing barrier in dialogue evaluation studies as conversational systems grow increasingly complex and multi-dimensional~\cite{tam2024framework}. However, most dialogue evaluation works focused on syntactical features, like fluency~\cite{ou-etal-2024-dialogbench}, grammar~\cite{lin2023llm}, and delivery accuracy  ~\cite{han2022measuring,chen2023automatic}, which ignores the the nature of dialogue functions--interactions or engagement.  In addition, these methods still lack interpretability and explanations of the evaluation results, rendering them unsuitable for deriving effective insights for conversation refinement or guidance for domain-specific conversations~\cite{de2023evaluation}, which leaves a gap for effectively evaluating the impact of deploying LLMs in domains like language education~\cite{long2024evaluating}. 

% Evaluating conversational dialogue quality while ensuring transparency and explainability remains a significant challenge.
Most existing dialogue evaluation methods focus on assessing the quality of machine-generated dialogues, emphasizing features like fluency~\cite{ou-etal-2024-dialogbench}, grammar~\citep{lin2023llm}, and  accuracy~\citep{han2022measuring,chen2023automatic}, while overlooking nuanced aspects such as interactions. These approaches primarily assess machine responses in isolation and often lack interpretability~\citep{smith2022human}. In contrast, traditional evaluations of human-to-human conversations rely heavily on manual coding and human interpretation~\citep{o2018use,mckenzie2000hope}, which, although detailed, are limited in scalability.

Recent studies have begun utilizing large language models (LLMs) to evaluate human dialogues, such as classroom interactions~\citep{long2024evaluating}. In the context of ESL conversation, \citet{gao2024cnima} proposed a fine-grained automatic evaluation tool that assesses dialogue quality on both micro (e.g., reference word usage) and macro levels (e.g., tone of utterance) for ESL speakers. Building on this foundation, our study adapts their framework to evaluate the quality of ESL dialogues. %under moderator intervention to identify effective moderation strategies.

% Recent works shift the focus from dialogue content accuracy to dialogue conversation natures that emphasis more on dialogue acts, or the nuances of conversational interaction. For example, Takehi et. al~\citet{takehi2023open} evaluates the dialogue quality based on turn-level dialogue acts, which makes the explanation rooted in dialogue turn-level. In addition, Gao et. al ~\citet{gao2024cnima} proposed an automatic evaluation tool to assess the dialogue quality on two levels, spanning from fundamental linguistic features (e.g., reference words, modal verbs, noun \& verb collocations and etc), to dialogue level organization, such as topic management, tone choice appropriateness, conversation opening and conversation closing with . These works serve as rare attempts in evaluating the dialogue quality based on the nature of conversations. Thus, the current study adopts the automatic dialogue quality evaluation tool from Gao et. al~\cite{gao2024cnima} to effectively measure the ESL dialogue quality under moderator actions, and to map the actual evaluation to explainable interpretations is crucial for providing insights in language education. 

\paragraph{Moderation in ESL group discussions}  

% A conversational moderator plays a pivotal role in discussions by (1) minimizing undesirable behaviors, (2) facilitating productive outcomes, and (3) ensuring balanced participation through various conversational interventions ~\citep{wright2009role, grimmelmann2015virtues}.

Previous studies have documented that moderators employ strategies such as linguistic scaffolding~\citep{kayi2013scaffolding, gagne2013cooperative}, providing instructions~\citep{hamzah2016effect}, seeking clarification, and offering acknowledgments~\citep{braham2009acknowledgement} to mitigate linguistic disparities~\citep{jones1999silence}, bridge cultural gaps~\citep{osman2007interaction}, and address knowledge deficiencies~\citep{asterhan2010online, vasodavan2020moderation}. However, these findings rely on manual evaluation methods such as interviews, case studies, and surveys~\citep{osman2007interaction, hew2008attracting, hamzah2016effect, kayi2013scaffolding}, limiting scalability and cross-context applicability.

Large-scale analyses of dialogue transcripts typically use dialogue acts to categorize speakers' intentions~\citep{d1985speech}. While existing moderation dialogue act schema (e.g., \citet{park2012facilitative}) provide a structured approach, they may not fully address the specific needs of ESL moderation. At the same time, defining entirely new dialogue acts in isolation could hinder cross-domain comparability. To address this challenge, we develop a tailored set of dialogue acts by adapting the WHoW moderation analysis framework. This approach ensures that the dialogue acts capture the nuances of ESL moderation while remaining compatible for broader cross-domain comparisons of moderator behavior.

In summary, there exists three major gaps: the absence of an automated method for measuring dialogue quality among ESL speakers in group discussions, the need for dialogue act schema specifically tailored for ESL group discussions moderation analysis, and the lack of quantification of the impact of conversation moderation in ESL settings.

\section{Dataset Development}\label{sec:dataset} 

Data sets of moderated, multi-party ESL discussions are scarce. To address this gap, we created the first corpus of ESL group discussions with paired moderated and non-moderated conditions. The structure and materials for the discussions come from an existing online ESL Conversation Club, designed to foster reflective and thought-provoking conversations for Asian ESL speakers that go beyond everyday small talk.
%\footnote{We  include the discussion guidelines for each session in Appendix~\ref{app:soulreme}.}

\subsection{ESL Conversation Session} \label{subsec:CSL-session}

Each session focuses on a central theme (e.g., “How AI impacts our daily life now and in the future”, Table~\ref{tab:topics_data}), accompanied by a background paragraph and three to five discussion questions (e.g., “Compared to AI, what are the things humans possess that are irreplaceable?”). All discussion materials are provided in Appendix~\ref{app:soulreme}. Sessions typically last 45 to 75 minutes and involve four to eight speakers online, and are moderated by experienced ESL moderators who have lived in English-speaking countries and have prior teaching experience. The moderator's responsibilities include (1) guiding participants in addressing the discussion questions, (2) fostering a friendly and engaging discussion environment, and (3) monitoring the discussion flow and timing. %Occasionally, the moderator also participates in the discussion to spark greater interest and engagement. 

\subsection{Data Collection} \label{subsec:CSL-collect}

\begin{table}[]
\begin{small}
\setlength{\tabcolsep}{0.3em}
% \resizebox{\columnwidth}{!}{%
\begin{tabular}{|L{0.49\linewidth}|M{1.2cm}|M{1.2cm}|M{1.2cm}|}
\hline
\multirow{2}{*}{Discussion Topic}  & mod & mod & non-mod  \\
&  ESL club & student & student \\ \hline
\textit{\textbf{Laugh}}: What makes you laugh? What does it convey and how is it perceived?  &    & \large{\checkmark}  & \large{\checkmark}   \\ \hline
\textit{\textbf{Romance}}: Romantic relationships in modern society.      & \large{\checkmark} & \large{\checkmark} & \large{\checkmark} \\ \hline
\textit{\textbf{Stress}}:  How to deal with stress?  & \large{\checkmark} & \large{\checkmark} & \large{\checkmark} \\ \hline
\textit{\textbf{Boss}}: How to cope with bosses with different cultural backgrounds. & \large{\checkmark}   & \large{\checkmark} & \large{\checkmark}   \\ \hline
\textit{\textbf{Time}}: Perspectives on how we perceive, manage, and value time.  & \large{\checkmark}  &   & \large{\checkmark} \\ \hline
\textit{\textbf{AI}}: How AI impacts our daily life now and the future?  &   & \large{\checkmark} & \large{\checkmark}\\ \hline
\textit{\textbf{Ghost}}: Are you superstitious? Why do people believe in `weird' things?  & & \large{\checkmark}    & \large{\checkmark}  \\ \hline
\end{tabular}%
\end{small}
% }
\caption{The seven discussion topics and their inclusion in different session settings: moderated (mod) online discussion club sessions, moderated student volunteer discussion sessions, and non-moderated student volunteer discussion sessions.}
\label{tab:topics_data}
\end{table}

The collected data were sourced from two formats: (1) participants of the online ESL conversation club and (2) international volunteer students. Data collection began with obtaining transcripts from the conversation club, hosted and recorded via Zoom, resulting in four sessions.\footnote{To preserve the original content and linguistic characteristics of ESL speakers, we do not edit the transcripts for grammar errors. Although common grammar errors (e.g., ``He don't like it.'') and incomplete sentences were present, manual review confirmed that the majority of the content was intelligible.} This primary dataset served as an initial field study to explore the general patterns and structure of the discussions.

We collected additional conversational data in a more controlled setting. In this setting: (a) the demographics of participants are more consistent; and (b) for each topic we ran two sessions, one with moderator and the other without (there's no overlap in terms of participants for the two sessions).\footnote{Some students attended only a single session, while others participated in multiple sessions. To ensure balanced exposure, students who joined multiple sessions were assigned to both moderated and non-moderated discussions. Note, however, that for the same topic, a student is only allowed to participate in either the moderated or non-moderated session.} All participants were native speakers of Chinese, most were postgraduate students or recent graduates who had been living in an English-speaking country for less than three years (IELTS scores of 6.5 to 7.5). Recruitment advertisements were distributed via social media platforms and the school's email channels. Participation was entirely voluntary and primarily motivated by participants' desire to practice English or engage in topic discussions. Consent form for data collection is provided in Appendix~\ref{app:consent_form}.

Our final data set (Table~\ref{tab:topics_data}) comprises four club sessions, and thirteen controlled discussion sessions (six moderated, seven non-moderated sessions).\footnote{We lost one moderated student volunteer session for the topic “Time” due to technical issue during recording.} Seven conversation topics were selected, all of which had been recently used in the conversation club. Four of these topics overlapped with the four earlier recorded sessions from the club. 
% Using these seven topics, we successfully established six parallel paired comparison groups, each discussing the same topics under moderated and non-moderated conditions. 
% In sum, while the combined data set provides us with sufficient data to study conversation quality overall, the subset of paired online discussion allows us to isolate the effects of moderation strategies from variations in discussion content.

We segmented the transcripts into shorter sections to facilitate our analysis. The nuanced linguistic structures and contextual dependencies in second-language conversations necessitated a manual approach to ensure contextually meaningful segmentation \cite{gao-etal-2025-interaction}. Two authors manually segmented each session based on sub-topic transitions within the main discussion theme, resulting in 6–11 segments per session. These segments served as the basis for dialogue interactivity quality evaluations and subsequent moderation dialogue act analyses. In total, we collected 16.5 hours of transcripts spanning 17 sessions. Descriptive statistics for the dataset are presented in Table~\ref{tab:data_stats}.

\begin{table}[]
% \resizebox{\columnwidth}{!}{%
\setlength{\tabcolsep}{0.48em}
\begin{small}
\begin{tabular}{@{}lc|cc|c@{}}
\toprule
Source        & Online club & \multicolumn{2}{l|}{Student   volunteer} & Total   \\ \midrule
Moderated     & \faCheckSquare         &  \faCheckSquare                 &  \faTimesCircle                 & ---    \\ \midrule
Sessions       & $4$           & $6$                   & $7$                  & $17$      \\ \midrule
Speakers Avg  & $6.5$         & $6$                   & $5.3$                & $44^{*}$  \\
Segments Avg  & $9.3$         & $9.7$                 & $7.9$                & $150$     \\
Sentences Avg & $836$         & $591$                 & $421$                & $9843$   \\
Mod Sent Avg  & $374$         & $298$                 & $0$                  & $3284$   \\
Tokens Avg    & $9450$       & $7364$               & $4595$              & $114152$ \\ \bottomrule
\end{tabular}%
\end{small}
% }
\caption{Descriptive statistics of the collected ESL group discussion transcripts, including the average number of speakers, segments, sentences, moderator sentences, and tokens per session type, and grand totals (right). $^*$The total value for speakers represents the number of unique individual participants.}
\label{tab:data_stats}
\end{table}

\section{Method} 
\begin{table*}[!t]
\centering
%\rowcolors{3}{gray!15}{white} % 设置较柔和的背景色
\begin{small}
\begin{tabular}{L{0.22\linewidth}L{0.17\linewidth}L{0.55\linewidth}}
\toprule
\textbf{Strategy}   & \textbf{Source}    & \textbf{Definition} \\ 
\midrule
 Information Probing & I $\&$ Probing   & Prompting participants to share thoughts, opinions, knowledge or experiences. \\
 \hline 
 Opinion Sharing & I $\&$ Supplement   & Express personal views, beliefs, or subjective opinions related to the topic. \\
 \hline 
Information Sharing & I $\&$ Supplement   & Provide factual, contextual content or knowledge to inform or orient others. \\
 \hline 
\multirow{2}{*}{Experience Sharing}  & \multirow{2}{*}{S $\&$ Supplement} & \multirow{2}{*}{Share a personal experience or anecdote.} \\ 
&  &  \\
 \hline 
Echoing  & I/S $\&$ supplement & Reinforce or support a prior statement by sharing similar views and thoughts. \\
 \hline 
Informational Interpretation & I $\&$ Interpretation & Interpret, clarify, reframe, summarize, paraphrase, or make connections to earlier conversation content. \\
 \hline 
Acknowledgement & S $\&$ Supplement & Recognize, validate, or show appreciation for another participant’s contribution, insight, or effort. \\
 \hline 
Backchanneling & S $\&$ Utility & Brief verbal or non-verbal responses for indicating active listening, understanding, or agreement. \\
 \hline 
\multirow{2}{*}{Social Utility}      & \multirow{2}{*}{S $\&$ utility}    & \multirow{2}{*}{Use polite or respectful phrases to show courtesy.} \\
&  &  \\
 \hline 
Coordinative Instruction & C $\&$ Instruction & Explicitly command, influence, or halt the immediate behavior of the recipients for coordinating the process of the session. \\
\bottomrule
\end{tabular}
\end{small}
\caption{ESLMOD strategies classes, along with their corresponding source WHoW motive labels (I for informational, S for social, and C for coordinative) and dialogue act labels and definitions.}
\label{tab:eslmod_def}
\end{table*}

To systematically assess the impact of moderation in ESL discussions, we first adapted the WHoW moderation  framework~\citep{chen2024whow} and applied topic modeling to our conversation transcripts to identify ten ESL-specific moderation strategies. Next, we incorporated an automated evaluation method for ESL group discussion quality based on \citet{gao2024cnima}. By comparing moderated and non-moderated sessions and analyzing the relationship between moderation strategies and dialogue quality, our approach quantifies the moderator’s impact in ESL discussions, providing a data-driven method to evaluate moderation effectiveness.

\begin{table*}[]
\centering
\begin{small}
\begin{tabular}{@{}llllllll@{}}
\toprule
Motive/Dialogue Act     & Probing        & Confronting      & Instruction       & Interpretation       & Supplement        & Utility         & Total \\ 
\midrule
Informational & $\mathbf{0.26(838)}$ & $0.01(17)$ & $0.00(11)$ & $\underline{0.09(290)}$ & $\mathbf{0.37(1223)}$ & $0.01(22)$ & $0.73(2401)$ \\
Coordinative & $0.01(37)$ & $0.00(2)$ & $\underline{0.04(119)}$ & $0.00(6)$ & $0.01(35)$ & $0.01(19)$ & $0.07(218)$ \\
Social       & $0.02(68)$ & $0.00(3)$ & $0.00(10)$ & $0.02(80)$ & $\mathbf{0.12(401)}$ & $\mathbf{0.10(315)}$ & $0.27(877)$ \\
\midrule
Total        & $0.26(856)$ & $0.01(19)$ & $0.04(124)$ & $0.10(342)$ & $0.46(1527)$ & $0.13(416)$ & $1(3284)$ \\
\bottomrule
% \textit{61.35}                         
\end{tabular}%
\end{small}
\caption{Probabilities (frequencies) of motives (rows) and dialogue acts (DA; columns), and conditional probabilities of DA given motive (cells), as identified using the WHoW framework. Bold values highlight intersected categories that surpass the threshold (0.1) for further domain adaption.}
\label{tab:whow_analysis}
\end{table*}

\subsection{Discovery of ESL Moderation Strategies} 
\label{subsec:discovery_method}

Conversational strategy analysis typically involves using dialogue acts—labels that represent speakers' intent—to identify sequential patterns within dialogue~\citep{chawla2022social}. In this study, rather than relying on existing dialogue acts developed for other contexts, we derived a set of ten ESL-specific moderation strategies (ESLMOD) by adapting the domain-agnostic WHoW moderation analytic framework~\citep{chen2024whow}. Definitions for the ESLMOD strategies are in Table \ref{tab:eslmod_def} and examples of each strategy in Appendix Table \ref{tab:slmoddef}.
 
The WHoW framework proposes a generic set of moderator-specific facilitation strategies, validated on moderated debates and panel discussions. It characterizes a moderator's role along three dimensions: Motives (“Why”), Dialogue Acts (“How”), and targeted speakers (“Who”). It defines three motive categories—informational, social, and coordinative—and six dialogue acts: probing, confronting, instruction, interpretation, supplement, and utility (see Appendix \ref{subapp:whow} for detailed definitions and examples). Following the WHoW framework, we used GPT-4o~\citep{openai2024api} to automatically annotate all moderator sentences in our data for these dimensions. We also prompted GPT-4o for a reason/justification of the predicted labels.\footnote{We manually validated the GPT-4o predictions of dialogue acts in $3$ out of the $10$ moderated sessions, achieving a macro-accuracy of $0.71$ and a macro-F1 score of $0.6$, both of which surpass the performance reported in the original study for debate and panel scenarios.}

Table \ref{tab:whow_analysis} presents the distribution of WHoW motives and dialogue acts, highlighting that the moderator's primary motivation is informational ($73\%$), with a secondary focus on fostering a social atmosphere and building relationships ($27\%$), and minimal emphasis on coordinating procedural rules or program-related aspects. In terms of functional roles (columns), moderators are heavily involved in supplementing information ($46\%$), driven by both informational ($37\%$) and social motives ($12\%$). Additionally, moderators frequently probe participants for input ($26\%$), interpret responses (10$\%$), and demonstrate active listening through various utility acts (13$\%$), such as back-channeling.

While WHoW could capture general moderation characteristics, its dialogue act categories are too broad to provide practical insights for ESL discussions. To refine these categories, we used K-means clustering to identify domain-specific dialogue acts~\citep{rus2012automated}. Based on initial WHoW predictions (Table \ref{tab:whow_analysis}), we applied frequency thresholds to identify the most dominant categories (12 categories with an intersecting probability $<2.5\%$ were removed). Of the remaining categories, four were selected for  adaptation, each accounting for $>10\%$ of instances. Two categories falling between $2.5\%$ and $10\%$ were included in the final list without further refinement.

For each selected prominent category, we extracted all sentences with the corresponding WhoW label, and the reasoning generated during the WHoW annotation which reflects the moderator’s intent. We then applied a separate BERTopic model~\citep{grootendorst2022bertopic} with k-means clustering to the generated reasonings (separately for each prominent category). The topic model identified fine-grained sub-categories for each original category. We validated our result using topic coherence, obtaining a refined set of domain-specific moderation strategies for ESL discussions.\footnote{Further details on the domain adaptation process and validation criteria are provided in Appendix \ref{app:topic_model}.}

After this process, we arrive at 10 ESL-specific moderation strategies (Table~\ref{tab:eslmod_def}). Specifically, the “Informational \& Supplement” category was subdivided based on the type of information shared (“Opinion Sharing”,  “Information Sharing”,  and “Echoing”). Similarly, the “Social \& Supplement” category was further divided into three subcategories: “Experience Sharing”, “Acknowledgment”, and “Echoing”. These subcategories are distinguished by the presence of agreement and the extent to which the discussion is expanded.\footnote{While “Acknowledgment” shows agreement or appreciation without expanding the discussion, “Experience Sharing” does not necessarily include explicit agreement, and “Echoing” both affirms the participant's contribution and elaborates on it.} We merged instances of “Echoing”—previously identified separately from the two intersecting categories—into a single strategy because it encompasses both social motives (relating to or agreeing with the participant) and informational motives (expanding the discussion). Additionally, the “Social Utility”  category was divided into “Backchanneling” (e.g., “Hmm”) and “Social Utility” (e.g., “Thank you!”).

With this refined set of strategies, we updated the WHoW prompt and re-annotated the moderator sentences (see Appendix Table \ref{tab:domain_prompt} for the revised prompt). To validate the new prompt, we randomly sampled 10 instances per refined category from the annotations generated by GPT-4o. Three PhD students, who were briefed on the strategy definitions and provided with examples, independently reviewed the annotations, either confirming the assigned label or selecting an alternative. This validation process yielded a macro-accuracy of $0.74$ and a Krippendorff's $\alpha$ of $0.41$, slightly exceeding WHoW's annotation performance (macro-accuracy = $0.71$) and demonstrating moderate reliability. To conclude, our adaptation process has produced 10 conversation moderation strategies tailored to ESL group discussion moderation analysis.

\subsection{ESL Multi-party Dialogue Evaluation}

% In this work, we adopt the dialogue evaluation framework of ~\cite{gao2024interaction}, which provides both micro-level and macro-level assessments of conversational quality. At the micro-level, the framework marks a total of 17 fundamental linguistics properties such as 'Code Switching`, 'Feedback in Next Turn`, and 'Negotiation of Meaning` across token and utterance levels. The macro-level examines the overall discourse, considering aspects like topic management, tone choice appropriateness, conversation opening and conversation closing. The dialogue level draw on a standardized rating scale from 1 to 5, where “1” indicates very unnatural and the overall dialogue utterances fail to achieve the communicative and interactive aim; and “5” means the speakers are able to employ richer linguistics resources (e.g., backchannels, collaborative finishes, speech acts) in dialogue communication to perform more authentic and natural interactions (e.g., more active and more natural in using a second language in dialogue communications). In addition, and independent evaluation metrics for quantifying the intercity and engagement score for dialogue overall quality as shown in Appendix~\ref{subapp:score15}. \citet{gao2024cnima} automate this framework by fine-tuning a series of models and deploying LLMs (e.g., GPT-4o) that accurately predict dialogue quality scores based on the micro-level and marco-level features from dialogues. Validation experiments show that the LLM automatic approach correlates strongly with human dialogue evaluations. 

We use the dialogue evaluation framework of~\citet{gao-etal-2025-interaction}, which provides: (1) micro-level; (2) macro-level; and (3) overall dialogue quality assessment. At the micro-level, the framework evaluates 17 fundamental linguistic properties, such as “Code Switching”, “Feedback in Next Turn”, and “Negotiation of Meaning”. At the macro-level, it assesses discourse quality through 4 dimensions: Topic Management, Tone Appropriateness, Conversation Openings, and Conversation Closings. They are rated on a standardized scale from 1 to 5, where 1 represents poor and
%which has limited linguistics resources and not naturally occurred in real-life conversations, and a score of 
and 5 high quality
%indicates better interactivity ability (e.g., backchannels, collaborative finishes, speech acts) to create fluent interactions in dialogue communication
(details in Appendix~\ref{subapp:score15}).
The overall dialogue quality in terms of conversation naturalness, the achievement of communication purposes and interactive participation engagements, was defined and evaluated by \citet{gao2024cnima} who automated and extended this framework using large language models (LLMs), showing that LLMs like GPT-4o can accurately predict overall dialogue quality, based on four macro-level quality scores (topic management, tone appropriateness, conversation opening/closing) and 17 micro-level features.
%JHL2: define overall dialogue quality RG3: addressed

%and 4 macro-level features, . They demonstrated that the automated LLM approach strongly correlates with human dialogue quality ratings. 
%JHL1: 1 and 5 represent what?
%Additionally, the framework includes independent metrics for interactivity and engagement, providing a comprehensive measure of overall dialogue quality 
%JHL1: I don't follow the previous line - I don't recall these independent metrics in the previous work

%JHL1: much better description than before, but I think it can be made a bit clearer. We should  start by describing there are three levels of assessment, micro-level, macro-level and overall dialogue score (define what they mean), and then describe how it was automated - using LLM to predict micro first, then use that output to predict macro, then use that output to predict the overall dialogue score.

As the dialogues in \citet{gao2024cnima} only have two speakers with similar proficiency, they computed one overall dialogue and 4 macro-level quality scores for each conversation. Given that our ESL conversations have \textit{multiple} speakers,
%\citet{gao2024cnima} focused on the overall quality assessment in the paired speaking context without much emphasis on individual speakers in the dialogue. Building on this,
we adapted their framework to compute an overall dialogue score and the macro-level quality scores for \textit{each speaker} in a conversation. The
%dialogue quality in multi-party discussions across four interactivity dimensions;
full prompt is given in Appendix~\ref{subapp:LLMeval}.

\section{Analysis and Results}\label{sec:experiment} 

In Section \ref{subsec:mod_analysis}, we characterize the ESL moderator behavior and role by examining the frequency distribution of the ESLMOD strategies. In Section \ref{subsec:dialogue_queality}, we assess the dialogue quality of group discussions and compare the outcomes of moderated versus non-moderated sessions. Finally in Section \ref{sec:effectivesta}, we identify most effective strategies for enhancing dialogue quality by analyzing the correlation between moderation strategies and dialogue quality. 

\subsection{Characterizing Conversational Moderation in ESL Group Discussion}
\label{subsec:mod_analysis}%

\begin{table}[t]
\centering
\begin{small}
\begin{tabular}{lccc}
\toprule
\textbf{Moderation Strategy} & \textbf{Frequency} & \textbf{\%}  \\
\midrule
Information Probing          & $849$ & $25.8\%$ \\
Information Interpretation   & $548$ & $16.7\%$ \\
Information Sharing          & $430$ & $13.1\%$ \\
Backchanneling               & $318$ & $9.7\%$ \\
Opinion Sharing              & $316$ & $9.6\%$ \\
Experience Sharing           & $247$ & $7.5\%$ \\
Acknowledgment               & $193$ & $5.9\%$ \\
Echoing                      & $179$ & $5.5\%$ \\
Coordination Instruction     & $146$ & $4.4\%$ \\
Social Utility               & $58$  & $1.8\%$ \\
\bottomrule
\end{tabular}
\end{small}
\caption{Frequency distribution of ESLMOD strategies in moderator sentences across online club and student discussion sessions.}
\label{tab:slmod_freq}
\end{table}

% Description and analysis of the SLMO distribution

Table \ref{tab:slmod_freq} presents the frequency distribution of ESLMOD strategies across all moderator sentences from the moderated sessions, including both online club and student discussion sessions. The data reveals that the moderator primarily employs strategies that encourage in-depth exploration and clarification of information, with the highest frequencies observed in “Information Probing”  (25.8\%) and “Information Interpretation”  (16.7\%). This pattern suggests a strong focus on eliciting detailed responses and ensuring that participants articulate their thoughts clearly. This is particularly beneficial in language learning contexts, as it helps reinforce comprehension, critical thinking, and language practice. Regarding the type of content shared by the moderator, “Information”  (13.1\%) and “Opinion”  (9.6\%) prevail over their own “Experience”  (7.5\%), indicating a greater emphasis on intellectual exchange rather than personal interaction in discussions. Strategies involving social support, such as “Acknowledgment” (5.9\%) and “Echoing” (5.5\%), are comparatively rare.

\subsection{Comparative Dialogue Quality Evaluation}
\label{subsec:dialogue_queality}% Rena Part: dialogue and four aspects also for speakers score 
%RG2: To Bryan, I run additional two samples t-tests to check the significance to make the story stronger 

\begin{table}[t]
\centering
\begin{small}
\setlength{\tabcolsep}{0.3em}

\begin{tabular}{cccccc}
\toprule
\multirow{2}{*}{\textbf{Moderated}} & \textbf{Dialogue} & \textbf{Topic} & \textbf{Tone} & \textbf{Conv} & \textbf{Conv} \\
& \textbf{Quality} & \textbf{Managmt} & \textbf{Choice} & \textbf{Open} & \textbf{Close} \\
\midrule
\faCheckSquare & {$\mathbf{4.14}$} & $\mathbf{4.03}$ & $3.48$ & $\mathbf{4.25}$ & $\mathbf{4.62}$ \\
\faTimesCircle & $3.41$ & $2.42$ & $2.68$ & $2.19$ & $2.07$ \\
\bottomrule
\end{tabular}
\end{small}
\caption{Dialogue quality scores for moderated (\faCheckSquare) and non-moderated sessions (\faTimesCircle). Significant differences under one-tailed t-test ($p{<}0.05$) are bolded.} \label{tab:rena_dialogue_score}
\end{table}

%RG3: a table here for speaker's score in different aspects. 
\begin{table}[t]
\centering
\begin{small}
\setlength{\tabcolsep}{0.4em}
\begin{tabular}{llllll}
\toprule
\textbf{Topic} &
  \textbf{\begin{tabular}[c]{@{}l@{}}Overall\\ Quality\end{tabular}} &
  \textbf{\begin{tabular}[c]{@{}l@{}}Topic\\ Managmt\end{tabular}} &
  \textbf{\begin{tabular}[c]{@{}l@{}}Tone \\ choice\end{tabular}} &
  \textbf{\begin{tabular}[c]{@{}l@{}}Conv\\ Open\end{tabular}} &
  \textbf{\begin{tabular}[c]{@{}l@{}}Conv\\ Close\end{tabular}} \\ \midrule
AI      & $1.09^\uparrow$ & $0.72^\uparrow$ & $0.28^\uparrow$ & $2.07^\uparrow$ & $1.74^\uparrow$ \\
Boss    & $2.01^\uparrow$ & $1.88^\uparrow$ & $0.90^\uparrow$ & $2.01^\uparrow$ & $1.72^\uparrow$ \\
Ghost   & $1.92^\uparrow$ & $1.72^\uparrow$ & $0.83^\uparrow$ & $2.71^\uparrow$ & $2.19^\uparrow$ \\
Laugh   & $1.21^\uparrow$ & $2.01^\uparrow$ & $0.07^\uparrow$ & $2.19^\uparrow$ & $2.01^\uparrow$ \\
Stress  & $1.09^\uparrow$ & $1.02^\uparrow$ & $0.28^\uparrow$ & $1.08^\uparrow$ & $0.87^\uparrow$ \\
Romance & $0.87^\uparrow$ & $0.93^\uparrow$ & $0.48^\uparrow$ & $1.29^\uparrow$ & $1.28^\uparrow$ \\\bottomrule
\end{tabular}
\end{small}

\caption{The difference in dialogue quality scores comparing moderated and non-moderated sessions within the same topic. ↑ denote improvement.}
\label{tab:interactivity-prediction-result}
\end{table}

\begin{table}[t]
\centering
\begin{small}
\setlength{\tabcolsep}{0.3em}
\begin{tabular}{cccccc}
\toprule
\multirow{2}{*}{\textbf{Moderated}} &
\textbf{Dialogue} & \textbf{Topic} & \textbf{Tone} & \textbf{Conv} & \textbf{Conv}\\ 
& \textbf{Quality} & \textbf{Managmt} & \textbf{Choice} & \textbf{Open} & \textbf{Close}\\
\midrule
\faCheckSquare & $\mathbf{3.60}$   & $\mathbf{4.30}$ & $3.19$ & $\mathbf{4.02}$  & $\mathbf{4.31}$\\ 
\faTimesCircle & $2.73$ & $2.86$  & $3.17$  & $2.80$ & $2.91$\\
\bottomrule
\end{tabular}
\end{small}
\caption{Dialogue quality scores for moderated (\faCheckSquare) and non-moderated (\faTimesCircle) sessions, when controlling for speakers. 
% for the same speakers who participated in both without moderator sessions and with moderator sessions, 
Significant differences under one-tailed t-test ($p{<}0.05$) are bolded.} \label{tab:rena_speaker_score}
\end{table}

%JHL1: the results are very positive, but the structure is a bit messy at the moment. Suggest:
%- first talk about table 5: explain what we are evaluating (overall dialogue + 4 macro scores), and then discuss the results
%-then talk about controlling for topic and speaker. one paragraph for topic (move table 7 up before table 6), and another paragraph for speaker
%-the current first paragraph seems a bit all over the place
%-the current second paragraph is mostly good, but I'd make it a bit more succinct; a lot of the discussions are somewhat fluffy and repetitive. really the key thing is that after controlled for topic we still see consistent improvement
%-the current third paragraph, again same comment. one thing we also need to do better is to explain how you control for speaker - the computation itself is not described 

To evaluate moderation’s impact on dialogue quality and interactivity, we analyze \textit{only} the controlled and paired student volunteering sessions (N=6). As shown in Table~\ref{tab:rena_dialogue_score}, the moderated sessions achieve a significantly higher overall dialogue quality score (4.14) than non-moderated ones (3.41). Conversation Opening and Closing significantly improve from 2.19 and 2.07 to 4.25 and 4.62, while Tone Choice shows only slight improvement. These results indicate that moderation enhances overall dialogue quality, particularly in structuring ESL multi-party discussions across all sessions with moderator interventions. To ensure the results in Table~\ref{tab:rena_dialogue_score} are not affected by topic or speaker differences, we conducted two additional analyses: (1) controlling for session topics and (2) controlling for speakers. When we controlled for topic (Table~\ref{tab:interactivity-prediction-result}), we saw consistent improvements for all scores across topics, even though the magnitude of the gain changes depending on the topic (``Romance'' e.g., seem to benefit least from the moderator's presence).
In line with Table~\ref{tab:rena_dialogue_score}, tone choice had the smallest improvement.
% To keep the setting consistent, we use the university student sessions with controlled settings (e.g., matched language proficiency, educational stage controlled, etc.) for dialogue interactivity quality evaluations and run the analysis based on these results for the impact of moderation dialogue acts. 

%First, we compared the overall dialogue quality of sessions with and without moderators, as summarized in Table~\ref{tab:rena_dialogue_score}. Next, we conducted a more detailed evaluation by examining four specific aspects of dialogue interactivity: \textbf{Topic Management, Tone Choice Appropriateness, Conversation Opening, and Conversation Closing.} for each segment within the overall dialogue sessions to make the evaluation more nuanced. Scores for each aspect were calculated for both moderated and non-moderated sessions, as shown in Table~\ref{tab:interactivity-prediction-result}. The results demonstrate consistent improvements in dialogue quality across all sessions with moderator interventions.

%, dialogue quality trends remained consistent, with statistically significant improvements across all dimensions except tone choice, aligning with Table~\ref{tab:rena_dialogue_score}.

Table~\ref{tab:rena_speaker_score} compares speakers who participated in both moderated and non-moderated sessions to ensure the variance of +moderator sessions and -moderator sessions are not due to participants individual differences, highlighting differences in dialogue quality across four interactivity aspects. Moderated sessions yield higher and more consistent scores, demonstrating the moderator's role in enhancing dialogue quality and reducing variability. In contrast, non-moderated sessions show greater score variability and lower averages, indicating that ESL speakers face significant challenges in managing discussions without external support.

\subsection{ESLMOD Strategies Comparison} \label{sec:effectivesta} 

\begin{table}[]
\begin{small}
\begin{tabular}{@{}llll@{}}
\toprule
Moderation Strategy    & Mean & Difference     & $p$-value \\ \midrule
Echoing      & $3.26$ & $0.51^{\uparrow\ast}$       & $0.04$ \\
Backchanneling             & $3.17$ & $0.77^\uparrow$             & $0.09$ \\
Experience Sharing         & $3.11$ & $0.12^\uparrow$             & $0.61$ \\
Coordination Instruction   & $3.09$ & $0.17^\uparrow$             & $0.46$ \\
Social Utility             & $3.09$ & $0.07^\uparrow$             & $0.79$ \\
Acknowledgement            & $3.06$ & $0.09^\uparrow$             & $0.77$ \\
Information Probing        & $3.02$ & $-1.23^\downarrow$          & $0.16$ \\
Information Interpretation & $3.01$ & $-0.33^\downarrow$          & $0.54$ \\
Information Sharing        & $2.95$ & $-0.54^\downarrow$          & $0.06$ \\
Opinion Sharing            & $2.89$ & $-0.68^{\downarrow\ast}$    & $0.03$ \\ \bottomrule
\end{tabular}

\end{small}
\caption{Mean dialogue quality scores for segments involving the specified ESLMOD strategy, alongside the score differences compared to segments without the strategy. Statistically significant differences (determined by a two-sample t-test, $p < 0.05$) are marked with an asterisk (*).}
\label{tab:slmod_act_eva}
\end{table}
%RG2: suggest putting Table 6 and 7 in one page, we can do this once all the content is done. 

To evaluate the effectiveness of ESLMOD strategies, we first computed an overall dialogue quality score $Q$ for each dialogue segment $D$. For each \(D\), each non-moderator speaker \(s_i\) is assigned a dialogue quality score \(q_i\) (ranging from 1 to 5), computed from Subsection \ref{subsec:dialogue_queality}. Let \(\mathcal{S}\) denote the set of non-moderator speakers in \(D\) and let \(t_i\) denote the number of tokens contributed by \(s_i\); we compute $Q$ by weighting the speaker scores \(q_i\):
\[
Q = \sum_{s_i \in \mathcal{S}} \frac{t_i}{\sum_{s_j \in \mathcal{S}} t_j} \cdot q_i
\]
This approach ensures that speakers with greater contributions have a proportionally larger influence on the segment’s quality score. Next, to assess the impact of a specific moderation strategy $m$, we computed the difference of $Q$ between segments with the strategy ($\mathcal{D}^{\text{m}}$) and those without ($\mathcal{D}^{\stkout{\text{m}}}$):
%\(\Delta Q^m\) between the mean \(Q_k\) of segments where the strategy was applied and the mean \(Q_k\) of segments where it was absent. Let \(\mathcal{D}_{\text{with}}^m\) and \(\mathcal{D}_{\text{without}}^m\) denote the sets of segments with and without the strategy $m$, respectively. Then:
\[
 \Delta_m = \frac{\sum_{D_k \in \mathcal{D}^{\text{m}}} Q_k}{|\mathcal{D}^{\text{m}}|}  - \frac{\sum_{D_k \in \mathcal{D}^{\stkout{\text{m}}}} Q_k}{|\mathcal{D}^{\stkout{\text{m}}}|} 
\]
We evaluate the statistical significance of using a two-sample t-test. Table \ref{tab:slmod_act_eva} presents the mean overall score ($Q$) for segments incorporating specific moderation strategies, along with difference ($\Delta_m)$ and $p$-value. Generally speaking, strategies involving the moderator’s sharing of content---such as information sharing, opinion sharing, and personal interpretations---tend to result in lower dialogue quality scores compared to segments where these strategies are absent. In contrast, strategies rooted in social motivation, including acknowledgment, social utility, experience sharing, echoing, and backchanneling, are with higher scores.

%“Echoing” exhibits a notably positive impact on dialogue quality, whereas “Opinion sharing” and “Information sharing” have significant negative effects. This suggests that moderators are most effective when they echo and recognize ESL speakers' contributions rather than offering additional information or opinions that might undermine their confidence. Moreover, a comparison between “Echoing” and “Acknowledgment” reveals that mere acknowledgment yields only minimal and insignificant improvement for ESL speakers; dialogue quality improves significantly only when moderators echo participants' contributions with supplementary supportive content. These findings align with prior educational studies~\citep{neusiedler2024engaging, mcclure2011we, gao2024listenership} that highlight the importance of recognization and positive minial backchanlling in enhancing ESL learners' participation.

When we focus on the most significant strategies (low $p$-value), the most effective moderator interventions are “Echoing” (e.g. “Yeah, absolutely, I feel like even just showing the willingness to to do or share housework can be applause.”) and “Backchanneling” (e.g. “Okay”). On the other hand, the least effective ones are “Opinion Sharing” and “Information Sharing”. This suggests that positive feedback is most helpful for ESL speakers as it helps build their confidence (and excessive information or opinion sharing might have the opposite effect). Moreover, comparing “Echoing” with simple “Acknowledgment” reveals that superficial positive feedback (e.g., “That is interesting.”) yields only minimal improvement, underscoring the need for engaging responses that extend the discussion. These findings align with prior educational studies~\citep{neusiedler2024engaging, mcclure2011we, gao2024listenership}, which emphasize the importance of recognition and positive engagement in enhancing ESL learners' participation.

%This suggests that moderators are most effective when reinforcing ESL speakers' contributions rather than introducing additional information or opinions, which may diminish speaker confidence. Furthermore, a comparison between “Echoing” and “Acknowledgment” reveals that simple acknowledgment has minimal impact, whereas echoing with supportive content leads to substantial dialogue improvements. These findings align with prior educational studies~\citep{neusiedler2024engaging, mcclure2011we, gao2024listenership}, emphasizing the role of recognition and positive minimal backchanneling in fostering ESL learner participation.

%RG3: To Bryan, I added some new content here to highlight the moderator strategry's results. 
%JHL1: if we look at p-value, backchanneling and information sharing is also quite significant (remember 0.05 is just an arbitrary threshold). My take is that echoing and backchanneling both look to have quite signicant positive impact, and information/opinion sharing negative impact. And interestingtly this is saying basically the best thing a moderator can do is to simply echo and acknowledge what ESL speakers say. When moderator try to share more information/opinion, contrary to expectation, it actually makes things worse! This is quite intuitive really, because what ESL speakers need is confidence boosting from the moderator, and echoing and backchanneling does basically that. Do prior education studies also find this? if so this is a very interesting finding, and it's something that we should highlight in the abstract and intro.

\section{Conclusions} \label{sec:conclusions}

In this study, We present the first ESL group discussion corpus, comprising 17 sessions (both moderated and non-moderated). Building on previous research in ESL dialogue evaluation~\citep{gao-etal-2025-interaction} and moderation analysis~\citep{chen2024whow}, we identify and integrate ten strategy classes with automated dialogue quality evaluation, allowing for the analysis of moderator influence without relying on external measures such as questionnaires. 
Our analysis shows that moderated sessions consistently exhibit a higher dialogue quality with less variability. Analysis of moderator behavior reveals a focus on informational exchange, but our comparative analysis suggests that socially motivated strategies improve dialogue quality. Notably, “Echoing” and “Backchanneling” are the most effective strategies, while “Opinion Sharing” and “Information Sharing” the least. These results show the value of social engagement and active listenership in ESL group discussions, offering important insights for moderator policy design.

All participants in this study were ESL speakers from similar cultural backgrounds. Previous research suggests that discussions involving a mix of ESL and native speakers, or those led by native-speaking moderators~\citep{zhu2001interaction, freiermuth2001native}, may produce significantly different dynamics. Future research could extend our approach to more diverse ESL dialogue settings, including interactions with native English speakers, to explore how cultural and linguistic diversity influences moderation effectiveness.

\section*{Limitations}\label{sec:limitations} %RG2: Bryan remember to check my words here to ensure the idea is accurate. 

This study has several limitations. First, the dataset is relatively small, which may limit the generalizability of the findings to broader contexts or diverse dialogue scenarios. Second, the analysis is based on sessions moderated by only two individuals, potentially introducing bias due to their specific moderation styles. And, it includes only language learners with an Asian background, and Chinese as native language. As a result, the findings may not fully capture the variability in moderation practices, nor their effects on dialogues involving participants from diverse or multiple cultures.

%Lastly, all participants in this study are ESL speakers. Based on previous research, we anticipate that discussions involving a mix of ESL and native speakers, or a native-speaking moderator~\citep{zhu2001interaction, freiermuth2001native}, could yield significantly different results. However, we leave this exploration for future studies. (RG3: I don't consider this as a limitation since our target is ESL discussion among ESL speakers which is different compared with other studies in nature.)

\section*{Ethics Statement}\label{sec:ethics}
This study was approved by the The University of Melbourne ethics board (Human Ethics Committee
LNR 1D), Reference Number 2022-24988-32929-3, and data acquisition and analysis has been taken
out to the according ethical standards.
% This study is conducted under the guidance of the ACL Code of Ethics. 
Personally identifyable information of all speakers as well as potentially offensive content was manually removed from the conversation transcripts.
% We manually filtered out potentially offensive content and removed all information related to the identification of all speakers. 

\normalem
\bibliography{custom}
\newpage
\onecolumn
\appendix

% \section{Appendix} \label{sec:appendix}

\section{Score definition of dialogue interactivity quality evaluation}

% Table~\ref{tab:dialoguefeaturelabel} shows the descriptions and definitions for dialogue's overall quality score.
\begin{table*}[!h]
\centering
\begin{tabular}{L{0.3\linewidth}L{0.6\linewidth}}
\hline
\textbf{\begin{tabular}[c]{@{}l@{}}Interactivity \\ Macro-level Features\end{tabular}} & \textbf{Definition} \\ \hline
Topic Management & The strategies and techniques used to control and navigate the flow of topics \\
Tone Choice Appropriateness & The suitability of the tone used in communication, ensuring it aligns with the context, audience, and purpose to convey the intended message  \\
Conversation Opening & The initial interaction or exchange that begins a dialogue, often setting the tone and context for the dialogue \\
Conversation Closing & The process of ending a dialogue or interaction, which involves signaling the conclusion of the discussion, summarizing key points, and often expressing a farewell \\
\hline
\end{tabular}
\caption{Definitions of macro-level interactivity features, with higher score emphasising on natural, authentic interaction and active engagement in the dialogue} \label{tab:dialoguefeaturelabel}
\end{table*}

\begin{table*}[ht]
\centering
\resizebox{1.0\textwidth}{!}{%
\begin{tabular}{lcl}
\toprule
\textbf{Interactivity Labels} & \textbf{Scores} &\textbf{Description of Scores} \\ 
\midrule
Topic Management &
\begin{tabular}[c]{@{}l@{}} 
  {[}5{]}\\  
  {[}4{]}\\
  {[}3{]}\\  
  {[}2{]}\\  
  {[}1{]}\\  
\end{tabular} &
\begin{tabular}[c]{@{}l@{}}
  topic extension with clear new context\\ 
  topic extension under the previous direction\\
  topic extension with the same content\\ 
  repeat and no topic extension\\ 
  no topic extension and stop the topic at this point\end{tabular}  \\ 
\midrule
Tone Appropriateness &
  \begin{tabular}[c]{@{}l@{}}
  {[}5{]}\\
  {[}4{]}\\
  {[}3{]}\\ 
  {[}2{]}\\ 
  {[}1{]}\\ 
  \end{tabular} &
  \begin{tabular}[c]{@{}l@{}}
  very informal \\ 
  quite informal, but some expressions are still formal \\ 
  relatively not formal, and most expressions are quite informal\\
  quite formal, and some expressions are not that formal\\
  very formal\end{tabular} \\ 
\midrule
Conversation Opening &
\begin{tabular}[c]{@{}l@{}}
  {[}5{]}\\ 
  {[}4{]}\\ 
  {[}3{]}\\ 
  {[}2{]}\\ 
  {[}1{]}\\ 
\end{tabular} &
\begin{tabular}[c]{@{}l@{}}
nice greeting and showing a good understanding of the opening of conversation in social interactions. \\ 
sounded greeting and showed a basic understanding of the social role. \\
general greeting but not understanding the social role well. \\ 
basic greeting.  \\ 
no opening, start the discussion immediately.
\end{tabular} \\ 
\midrule
Conversation Closing &
\begin{tabular}[c]{@{}l@{}}
{[}5{]}\\ 
{[}4{]}\\ 
{[}3{]}\\ 
{[}2{]}\\ 
{[}1{]}\\ 
\end{tabular} &
\begin{tabular}[c]{@{}l@{}} detailed summarization and smooth transition to the closing of the conversation. \\ 
transit to the closing naturally, but without summarising the discussion. \\
transit to the discussion. \\ 
demonstrate a translation to the end of the conversation. \\ 
no closing, directly stop the conversation.
\end{tabular} \\ 
\bottomrule
\end{tabular}
}
\caption{Description of scores for dialogue-level interactivity labels. Higher score indicates better interactivity ability, for example, \textit{Tone Appropriateness} scores higher with more informality shows that the speakers are able to employ more active linguistics resources in dialogue communication to perform more causal and natural interactions compared with the formal tone, which has limited linguistics resources and not naturally occurred in real-life conversations.}
\label{tab:dialogue-feature-label}
\end{table*}

\newpage

\section{Overall Dialogue Quality Score Description and Definitions}\label{subapp:score15}
The following Table 12 shows the descriptions and definitions for dialogue's overall quality score.
\begin{table*}[h]
\centering
\begin{tabular}{M{1.5cm}L{0.7\linewidth}}
\hline
\textbf{Scores} & \textbf{Descriptions} \\ \hline
5 & Smooth and fluent daily communication, easy and pleasant through the whole chat \\
4 & Somewhat less fluent communication, but the communication purpose is achieved  \\
3 & Slightly awkward communication in some places, such as not being able to understand the other person’s question \\
2 & Overall communication is not fluent and mostly awkward, but some parts can be mutually understood \\
1 & Unable to accurately achieve the communication purpose, awkward conversation, and failed to talk throughout the conversation.\\ \hline
\end{tabular}
\caption{Score description for overall dialogue quality} 
\label{tab:score15}
\end{table*}

\textbf{Prompts for GPT-4o Dialogue Overall Evaluation} \label{subapp:LLMeval} 

The following Table 11 shows the prompts for dialogue overall quality score with GPT-4o.

\begin{table}[!ht]
\centering
\begin{tabular}{@{}p{4cm}p{10cm}@{}}
\toprule
\textbf{Field} & \textbf{Description} \\ 
\midrule
\textbf{Conversation} & A dialogue of second language Chinese conversation. \\
\hline 
\textbf{Output Fields} &  \textbf{Score}: The score of the interactivity of the English second language dialogue (1 to 5). \\ 
& \textbf{Rationale}: The reason why and how the score is made based on each participant's utterance. \\
\hline 
\textbf{Evaluation Criteria} & 
\textbf{5}: Smooth and fluent daily communication, easy and pleasant. \\
& \textbf{4}: Somewhat less fluent communication, but the communication purpose is achieved. \\
& \textbf{3}: Slightly awkward communication, such as not being able to immediately understand the other person's question with hesitation. \\
& \textbf{2}: Overall communication is not fluent and awkward, but some parts can be mutually understood. \\
& \textbf{1}: Unable to accurately achieve the communication purpose, awkward conversation, failed to talk throughout the conversation. \\
\toprule
\textbf{Human Validation} & Authors from this study verified the LLM results manually and pass the 75\% judgement compared with human evaluation. \\ 
\bottomrule
\end{tabular}
\label{tab:evaluation_task}
\caption{LLM Dialogue Overall Dialogue Quality Evaluation Prompts and Human Validation Process} 
\end{table}

\newpage

\section{WHoW Analytic Framework}\label{subapp:whow}

% \nopagebreak[4]
% \vspace*{-5px}
\begin{table}[H]
\resizebox{\textwidth}{!}{
\begin{tabular}{p{0.115\linewidth} p{0.185\linewidth} p{0.785\linewidth}}
\toprule
{\bf Dimension} & {\bf Label} & {\bf Definition} \\ \midrule
\multirow{6}{*}{Motives} & Informational (IM) & Provide or acquire relevant information to constructively advance the topic or goal of the conversation. \\
 & Coordinative (CM) & Ensure adherence to rules, plans, and broader contextual constraints, such as time and environment. \\
 & Social (SM) & Enhance the social atmosphere and connections among participants by addressing feelings, emotions, and interpersonal dynamics within the group. \\ \midrule
\multirow{11}{2cm}{Dialogue acts} & Probing (Prob) & Prompt speaker for responses.\\  \\
 & Confronting (Conf) & Prompt one speaker to response or engage with another speaker's statement, question or opinion. \\
 
 & Instruction (Inst) & Explicitly command, influence, halt, or shape the immediate behavior of the recipients. \\
 & Interpretation (Inte) & Clarify, reframe, summarize, paraphrase, or make connection to earlier conversation content. \\
 & Supplement (Supp) & Enrich the conversation by supplementing  details or information without immediately changing the target speaker's behavior. \\
 & Utility (Util) & All other unspecified acts. \\ \midrule
Target Speaker & Target speaker (TS) & The group or person addressed by the moderator. \\ \bottomrule
\end{tabular}
}
\caption{Definitions and acronyms for the labels across the three dimensions: motives (Why), dialogue acts (How), and target speakers (Who). Target Speaker is a categorical variable with values corresponding to each participant in the dialogue, plus ``audience'', ``self'', ``everyone'', ``support side'', ``against side'', ``all speakers'', and ``unknkown''.}
\label{tab:whowdef}
\end{table}
\newpage
% \nopagebreak[4]
\begin{table}[H]
\centering
\resizebox{\columnwidth}{!}{%
\begin{tabular}{|p{0.05\linewidth} | p{0.35\linewidth} p{0.35\linewidth} p{0.35\linewidth}|}
\toprule
DAs & IM & CM & SM \\ \midrule
\rowcolor[HTML]{EFEFEF} 
Prob & Can you take that on? (prompting)\newline As long as the political spectrum is covered overall, what’s wrong with that? (follow up question)\newline Siva? (name   calling prompt) & Which of you would like to go first? (preference inquiry)\newline Did this gentleman come down yet? (coordinative question)\newline It's working, right? (question managing environment) & Is that a relief to you or-- (asking feeling)\newline Could you tell us your name, please? (social question)\newline Do you have eyeglasses? (humour question) \\ \midrule
Conf & That landed pretty well I think, so can you respond to that? (counter confronting)\newline On this side, do you want to respond, or do you agree? (consensus confronting)\newline You actually asked a perfect question, and so Mark Zandi, do you want to take that on? (confronting question) & The other side care to respond, if not I’ll move on.(coordinative consensus)\newline Response from the other side, or do you want to pass? (coordinative confronting)\newline Marc Thiessen, do you want to join your partner on this one, because I think-- (coordinative consensus) & Bryan Caplan, I think he just described your fantasy, come true.(social confronting)\newline I'd love to hear your answer to that question, so go for it. (confronting with affective appeal)\newline Jared Bernstein, the guy you called “nuts” just said you're unfair. (humour confronting) \\ \midrule
\rowcolor[HTML]{EFEFEF} 
Inst & Can you frame your question as a question? (articulate instruction)\newline Relate that point to this motion. (back to topic)\newline I want to stay on the   merits of the Obama plan. (manage topic) & Remember, about 30 seconds is what you'll get. (time control)\newline Can you go up three steps, please, and turn right? (coordinating instruction)\newline I'll be right back after this message. (program management) & Do not be afraid. (emotion instruction)\newline Those who agree, just a round of applause to that. (pro-social instruction)\newline --because it's turning into a personal attack. (stop anti-social) \\ \midrule
Inte & So, Matt, you're saying that it's not true that it's inevitable that Amazon will control everything. (summarization)\newline Their point is that it would be a bad thing. (simplification)\newline But that would be the question of mobility. (reframe) & That was an ambiguous signal. (situation interpretation)\newline You're pointing to Lawrence Korb.(preference interpretation)\newline And you want the side arguing   for the motion to address that (preference interpretation) & I think it was a rhetorical question, and it got a good laugh. (humour interpretation)\newline And it's a little bit insulting almost to say (toxicity interpretation)\newline —honestly, I don’t think that was an—a personal attack— (toxicity interpretation) \\ \midrule
\rowcolor[HTML]{EFEFEF} 
Supp & I agree that it is.(agreement)\newline The fact is that one of the US manufacturers, with 1 percent of its yearly production, would run us out of the whole market.(add information)\newline They had never paid any attention whatsoever to Africa. (share opinion) & Fifty-one of you voted against the motion. (vote reporting)\newline And the mic’s coming down to you. (describe situation)\newline Round two is where the debaters   address each other directly (rule explanation) & You have a colorful sleeve. (social chit-chat)\newline I hate to reward it but I'm going to. (encouragement)\newline And I think all of us probably share a sense that we want things to improve. (state common feeling) \\ \midrule
Util & Fair question. (acknowledgement)\newline Right (acknowledgement)\newline  So the-- (floor grabbing) & All right. (backchanneling)\newline Actually, I-- (floor grabbing)\newline Well—(floor   grabbing) & Thank you Evgeny Morozov. (thanks)\newline I'm sorry. (apology)\newline Hi. (greeting) \\ \bottomrule
\end{tabular}%
}
\caption{This table presents a collection of exemplar sentences from the original paper at the intersection of the motives and dialogue acts dimensions.}
\label{tab:example_sents}
\end{table}

\begin{table}[]
\resizebox{\textwidth}{!}{%
\begin{tabular}{|p{0.170\linewidth} | p{\linewidth}|}
\toprule
Section & Prompt Part \\ \midrule
\rowcolor[HTML]{EFEFEF} 
Role \& Topic & Your role is an annotator, annotating the moderation behavior and speech of a debate TV show. The debate topic is ``When It Comes To Politics, The Internet Is Closing Our Minds"\newline \\
Task Instruction & given the definition and the examples, the context of prior and posterior dialogue, please label if the target utterance carries informational motive?\newline \\
\rowcolor[HTML]{EFEFEF} 
Dimension Instruction & Motives: Motives are the high level motivation that the moderator aim to achieve. The definitions and examples of the informational motive are below:\newline\\
Label definition & informational motive: Provide or acquire relevant information to constructively advance the topic or goal of the conversation.\newline \\
\rowcolor[HTML]{EFEFEF} 
Label Examples & examples: “Why do you think minimum wage is unfair?” (Relevant information seeking.) “The legal system has many loopholes.” (Expressing opinion.) “Yea! I agree with your point!” (Agreement relevant to the topic.)  “The law was established in 1998.” (Providing topic relevant information.)\newline \\
Dialogue Prior Context & Dialogue context before the target sentence:\newline\newline (including dialogue up to 5 utterance prior)\newline\newline  Eli Pariser (for): Right, and the question is, can you trust them?\newline\newline  John Donvan (mod): Let me-- Jacob, I think Eli left a pretty good image hanging out there, of these folks truly not knowing how much they don’t know and believing what they’re getting and not understanding how slanted it is.\newline \\
\rowcolor[HTML]{EFEFEF} 
Target Sentence & Target sentence:\newline\newline John Donvan (mod): That landed pretty well I think, so can you respond to that?\newline \\
Dialogue Post Context & Dialogue context after the target sentence:\newline\newline Jacob Weisberg (against): But a guy who called into a radio show? I know the plural of anecdote is data......(more)\newline\newline  John Donvan (mod): Siva. \newline\newline (including dialogue up to 2 utterance after the target.)\newline\\
\rowcolor[HTML]{EFEFEF} 
Formatting Instruction & Please answer only for the target sentence with the JSON format:\{"motives": List(None or more from "informational motive", "social motive", "coordinative motive"),"dialogue act": String(one option from "Probing", "Confronting", "Supplement", "Interpretation", "Instruction", "All Utility"),"target speaker(s)": String(one option from "0 (Unknown)", "1 (Self)", "2 (Everyone)", "3 (Audience)", "4 (Eli Pariser- for)", "5 (Siva Vaidhyanathan- for)", "6 (Evgeny Morozov- against)", "7 (Jacob Weisberg- against)", "8 (Support team)", "9 (Against team)", "10 (All speakers)"),"reason": String\}\newline\newline For example: answer: \{"motive": {[}"informational motive"{]}, "dialogue act": "Probing",  "target speaker(s)": "7 (Joe Smith- for)", "reason": "The moderator asks a question to Joe Smith aimed at eliciting his viewpoint or reaction to a statement from the recent policy change for combatting climate change......"\}\newline \\ \bottomrule
\end{tabular}%
}
\caption{A simplified WHoW framework prompt for annotating a target sentence regarding the motives, dialogue acts, and the target speaker. The 'Dimension Instruction' section is repeated for the 'Dialogue Act' dimension, while 'Label Definition' and 'Label Examples' are repeated for each label under the motives and dialogue acts dimensions.}
\label{tab:whow_prompt}
\end{table}

\section{ESLMOD moderation strategies}\label{subapp:eslmod}

\begin{table}[H]
\resizebox{\columnwidth}{!}{%
\begin{tabular}{|p{0.3\linewidth}|p{0.75\linewidth}|}
\hline
\textbf{Category} &
  \textbf{Examples} \\ \hline
\textbf{Information Probing} &
  “Anyone else who wants to share their thoughts or opinions about what the purpose of a relationship is for them?” “Related to stress. And how do you manage in such situations?” “Yeah. How about you, Chantelle?” “Do you agree with this statement?” \\ \hline
\textbf{Opinion Sharing} &
  “For me, managing stress is all about maintaining a good work-life balance.” “To me, sharing housework equally is a sign of respect and partnership in a relationship.” “I believe that having diverse perspectives in a team leads to more creative solutions.” \\ \hline
\textbf{Information Sharing} &
  “Today’s topic is related to the recent trends in the job market.” “Research shows that group discussions can improve second language acquisition by increasing practice opportunities.” “You can find free online courses on platforms like Coursera or edX to learn new skills.” \\ \hline
\textbf{Echoing} &
  “Yeah, my friend had a similar experience when he was in the US, as he struggled to find a job.” “I can relate to that feeling of being overwhelmed when learning a new language. I have been through it too.” “I completely agree—my friend also struggled with finding a balance between work and family responsibilities.” \\ \hline
\textbf{Experience Sharing} &
  “There was a time when I had to make a tough decision about changing my career path—it was such a challenging moment for me.” “Once, during my university days, I stayed up all night preparing for a group project because I wanted everything to be perfect.” \\ \hline
\textbf{Acknowledgement} &
  “That is a very interesting insight.” “Great point, and I think it really ties back to what we were discussing earlier.” “I appreciate you bringing this up—it’s a really valuable perspective.” “Thanks for sharing that example—it really helped clarify the idea.” \\ \hline
\textbf{Backchanneling} &
  “Yeah.” “Hmm.” “Okay.” “Uh-huh.” “Right.” “I see.” “Mhm.” \\ \hline
\textbf{Social Utility} &
  “Goodbye!” “Thank you!” “Please, go ahead!” “Excuse me.” “I appreciate your time.” \\ \hline
\textbf{Informational Interpretation} &
  “If I understand correctly, you’re suggesting that online courses are beneficial because they provide flexibility.” “To summarize, the main takeaway here is that building relationships in the workplace helps reduce stress.” “In other words, you’re arguing that peer feedback plays a critical role in language learning success.” \\ \hline
\textbf{Coordinative Instruction} &
  “Can we wrap up this discussion and move on to the next point?” “I’d like everyone to think about this question and share your thoughts one by one.” “Now everyone is here, let’s start the session.” “Please turn off your microphone when you are not speaking.” \\ \hline
\end{tabular}%
}
\caption{ESLMOD strategies categories and examples.}
\label{tab:slmoddef}
\end{table}

\newpage

\begin{table}[H]
\resizebox{\textwidth}{!}{%
\begin{tabular}{|p{0.170\linewidth} | p{\linewidth}|}
\toprule
Section & Prompt Part \\ \midrule
\rowcolor[HTML]{EFEFEF} 
Role \& Topic & Your role is an annotator, annotating the moderation behavior of a second language speakers" English conversation session. The topic is "Are you superstitious? Why do people believe in ‘weird’ things?"\newline \\
Task instruction & given the definition and the examples, the context of prior and posterior dialogue, please label which dialogue act the target sentence belong to? And who is the moderator talking to?\newline \\
\rowcolor[HTML]{EFEFEF} 
Dimension Instruction & Dialogue act: Dialogue acts is referring to the function of a piece of a speech/sentence. The definitions and examples of the dialogue acts are below:\newline\\
Label Definition & Information Probing: Prompting participants to share their thoughts, opinions, or experiences by posing questions or directly inviting input from individuals or the group.\newline
\\
\rowcolor[HTML]{EFEFEF} 
Label Examples & Examples: “Anyone else who want to share their thought or opinion about why they what is the purpose of relationship for them?” “Related to stress. And how do you manage in such situation?” “Yeah. How about you, Chantelle?” “Anyone else who want to share their thought or opinion about why they what is the purpose of relationship for them?” “Do you agree with this statement?”\newline \\
Dialogue Prior Context & Dialogue context before the target sentence:\newline(including dialogue up to 5 utterance prior)\newline\newline  Andrew (participant): And why should I contribute a huge amount of effort in my life to these gods, to me they don't exist, or maybe they do.\newline\newline
Andrew (Participant): It's just that they gotta find a way to manifest themselves at least.\newline\newline
Andrew (Participant): But I do believe in luck in and also, I believe in all the coincidences in life, like, Yeah.\newline\newline 
Bryan (Host) (Moderator): But I think it's not probably not just about religions.\newline\\
\rowcolor[HTML]{EFEFEF} 
Target Sentence & Target Sentence:\newline\newline Bryan (moderator): But sometimes they people just have this kind of like superstitution, like, you know if today, by wearing this color, or I have my watch, which is, I don't know, and it might probably bring luck to me. \newline\\
Dialogue post context & Dialogue context after the target sentence:\newline\newline Christine (participant): Yeah, I think, like, actually, I think, like, everyone, have some sort of religious or cultural specifications like at least influenced by one......(more)\newline\newline (including dialogue up to 2 utterance after the target.)\newline\\
\rowcolor[HTML]{EFEFEF} 
Formatting Instruction & Please answer only for the target sentence with the JSON format:{"dialogue act": String(one option from 0 (Information Probing), 1 (Opinion Sharing), 2 (Information Sharing), 3 (Echoing), 4 (Experience Sharing), 5 (Acknowledgement), 6 (Backchanneling), 7 (Social Utility), 8 (Information Interpretation), 9 (Coordination Instruction)),"target speaker(s)": String(one option from "0 (Unknown)", "1 (Everyone)", "2 (Emma)", "3 (Andrew)", "4 (Christine)", "5 (Jodie)", "6 (Leo)", "7 (Yuki)", "8 (Yale)"),"reason": String}\newline\newline For example: answer: \{"dialogue act": "1 (Information Probing)", "target speaker(s)": "3 (Joe Smith)", "reason": "The moderator asks a question to Joe Smith aimed at eliciting his viewpoint or reaction to a statement from the recent policy change for combatting climate change......"\} \newline\\ \bottomrule
\end{tabular}%
}
\caption{An example of ESLMOD prompt updated from the WHoW prompt incorporating the identified moderation strategies for ESL group discussions. The 'Label Definition' and 'Label Examples' subsections are repeated for each newly identified moderation strategy.}
\label{tab:domain_prompt}
\end{table}

\section{Domain Adaption Using Topic Modeling with WHoW}
\label{app:topic_model}

\begin{table}[H]
\resizebox{\textwidth}{!}{%
\begin{tabular}{|p{0.25\linewidth} | p{0.2\linewidth} |p{0.1\linewidth}| p{0.25\linewidth}| p{0.1\linewidth} |p{0.15\linewidth}| p{0.2\linewidth}|}
\hline
motive / dialogue act &
  hyper-parameters &
  coherence score &
  topic models words &
  cluster size &
  decision &
  refined lable \\ \hline
\multirow{3}{*}{Informational probing} &
  \multirow{3}{0.2\linewidth}{kmean\_n\_clusters=3 umap\_n\_neighbor=5, umap\_min\_dists=0.0} &
  \multirow{3}{*}{0.729} &
  question, asking, ask, aim, elicit &
  574 &
  \multirow{3}{*}{merge} &
  \multirow{3}{*}{Informational probing} \\ \cline{4-5}
 &
   &
   &
  ask, address, question, attempting, information  &
  221 &
   &
   \\ \cline{4-5}
 &
   &
   &
  asking, information, perspective, question, seek &
  70 &
   &
   \\ \hline
\multirow{3}{*}{Informational supplement} &
  \multirow{3}{0.2\linewidth}{kmean\_n\_clusters=3, umap\_n\_neighbor=10, umap\_min\_dists=0.5} &
  \multirow{3}{*}{0.453} &
  additional, opinion, personal, perspective, insight &
  823 &
  expand &
  Opinion sharing \\ \cline{4-7} 
 &
   &
   &
  information, additional, provides, response, supplement &
  292 &
  expand &
  Information sharing \\ \cline{4-7} 
 &
   &
   &
  add, build, reflect, reinforce, agree &
  82 &
  expand &
  Echoing \\ \hline
\multirow{4}{*}{Social supplement} &
  \multirow{4}{0.2\linewidth}{kmean\_n\_clusters=4, umap\_n\_neighbor=30, umap\_min\_dists=0.1} &
  \multirow{4}{*}{0.474} &
  personal, shares, experience, atmosphere, social &
  213 &
  expand &
  Experience Sharing \\ \cline{4-7} 
 &
   &
   &
  expresses, positive, appreciating, agreement, statement &
  52 &
  \multirow{2}{*}{merge} &
  \multirow{2}{*}{Acknowledgement} \\ \cline{4-5}
 &
   &
   &
  acknowledges, point, gratitude, previous, statement &
  74 &
   &
   \\ \cline{4-7} 
 &
   &
   &
  respond, compliment, reflect, serve, statement &
  50 &
  merge to Echoing &
  Echoing \\ \hline
\multirow{2}{*}{social utility} &
  \multirow{2}{0.2\linewidth}{kmean\_n\_clusters=2, umap\_n\_neighbor=10, umap\_min\_dists=0.0} &
  \multirow{2}{*}{0.453} &
  expresses, acknowledges, gratitude, uses, bye &
  73 &
  expand &
  Social utility \\ \cline{4-7} 
 &
   &
   &
  serves, acts, serving, backchannelling, respond &
  249 &
  extend &
  Backchanneling \\ \hline
\end{tabular}%
}
\caption{This table presents the optimized hyper-parameters for BERTopic applied to the four prominent WHoW intersected labels. It details the $k$ value used for KMeans clustering, as well as the number of neighbors and the minimum distance parameters for UMAP. Additionally, the table reports the coherence score for each cluster and lists the top five keywords for the sub-topic clusters. The final two columns indicate whether each cluster was merged or expanded, along with the manually refined names for the new sub-topics.}
\label{tab:topicmodel}
\end{table}

\twocolumn

\subsection{Motivation and Advantage of Using WHoW as a Foundational Framework}

The WHoW analysis provides a high-level characterization of the moderator's role in specific conversational scenarios. For example, a debate moderator tends to prioritize information dissemination and coordination, often adopting a stronger functional role in managing conflicts and providing instructions. While these general insights allow for comparisons across scenarios, they may lack the granularity necessary to guide specific strategies or actions. For instance, identifying a moderator as high in information supplement indicates that information is being shared but does not specify the type of information (e.g., opinions or experiences).

Nevertheless, the WHoW analysis serves as a foundational framework—a “skeleton”—that can be refined into a more detailed set of strategies tailored to specific domains. Our literature review indicates that dialogue act schemas developed for various domains often include fine-grained categories for domain-specific acts, while relying on coarser categories for more general or less frequent acts. For example, in e-rulemaking discussions~\citep{park2012facilitative}, probing interventions are subdivided into three types: prompting users to provide additional information, encouraging them to propose or consider solutions, and posing open-ended questions. Similarly, information supplements are categorized into providing details about proposed rules, pointing to relevant resources, and identifying characteristics of effective commenting. However, only a single type of intervention for information interpretation is included: correcting misstatements. Conversely, in e-learning scenarios~\citep{vasodavan2020moderation}, the focus shifts, with three distinct types of interpretive interventions: summarizing discussions, highlighting contributions, and archiving information. The WHoW analysis identifies broad, high-frequency categories that can serve as a starting point for further refinement into fine-grained, domain-specific dialogue acts.

A key advantage of using WHoW as the foundational framework for domain adaptation is that it allows for easier comparison of moderator behavior across different domains. For example, the schemas developed in the two earlier studies are not directly comparable because they use different label sets. However, if the labels were based on the WHoW framework, it would be possible to measure the similarity between moderators' functions and motivations in various scenarios. This kind of cross-domain comparison can help reveal patterns that are consistent across contexts, offering useful insights into effective moderation practices and potentially guiding improvements in moderation strategies across different settings.

\subsection{Dialogue Act Domain Adaption For ESL Discussion Moderation}

\subsubsection{ Labels Selection}

Building on the WHoW analysis applied to the ESL moderated conversation datasets, we excluded 12 of the 18 intersected motive/dialogue categories that accounted for less than 2.5\% of all instances. Among the remaining combinations, four prominent categories emerged—informational probing, informational supplement, social supplement, and social utility—each representing more than 10\% of instances. These labels then served as the basis for further exploration and potential domain-specific expansion. Two categories that fell between 2.5\% and 10\% were directly included in the final list without additional exploration.

It is important to note that the selected thresholds were tailored for the current study. Categories representing less than 2.5\% (approximately 75 samples) were deemed too insignificant, while those with fewer than 10\% (around 300 samples) were considered insufficient for robust topic modeling. These thresholds can be adjusted based on the specific use case and overall sample size.

\subsubsection{Pre-processing annotation reasons}

For each prominent label, we extracted the associated moderator sentences and applied topic modeling to explore potential specifications. To ensure that the topic modeling results reflected the moderator’s intent and actions rather than merely the discussion topics, we leveraged the “reasons” generated by GPT-4o during the WHoW analysis. For example, GPT-4o generated the following reason:

\begin{displayquote}
``The moderator is providing a contextual or explanatory statement intended to set the scene for the discussion about societal and cultural beliefs in things beyond rational or logical understanding, which is shared information intended for all participants.''
\end{displayquote}

Typically, the first sentence of a reason summarizes the core intent and action of the moderation, while subsequent sentences elaborate on motives, dialogue acts, and target speakers. To make the topic modeling results more sensitive to the moderator's intent, we applied the following pre-processing steps:

\begin{itemize} \item Extract the first sentence from each reason generated by GPT-4o during the WHoW analysis. \item Process the sentence with SpaCy’s dependency parser to identify the root verb and its direct object subtree. For instance, the sentence “The moderator shares a personal anecdote about his experience working in an AI company to contribute work-related insights to the topic 'The impact of AI'” is reduced to “shares a personal anecdote about his experience working.” \item Curate an additional list of stop words, including speaker names (e.g., “Amy”) and keywords specific to the discussion topic, such as “AI” and “stress.” \end{itemize}

\subsubsection{Topic modeling for identifying domain specific dialogue acts}

We applied BERTopic~\citep{grootendorst2022bertopic}, a widely used neural topic modeling approach that leverages pre-trained BERT embeddings, combined with K-means clustering. We optimized the number of clusters ($k$) within a range of 2 to 5 to identify potential sub-categories. We selected K-means over DBScan to maintain control over the number of clusters. Additionally, we employed U-MAP for dimensionality reduction of the sentence representation vectors, reducing the impact of small sample sizes and sparse vectors on the clustering process. For each prominent intersecting label, we ultimately identified the optimized hyper-parameter sets, as summarized in Table~\ref{tab:topicmodel}.

Decisions on whether to expand or merge sub-topics were based on the following criteria:

\begin{itemize} \item Do the sub-topic's top five representative keywords distinctly differ from those of other acts or WHoW dialogue acts? \item Are the original reasons associated with the sub-topic's samples clearly distinguishable from those of other acts or WHoW dialogue acts? \item Do the sub-topic's keywords form a coherent theme? \item Are the moderator sentences in the sub-topic notably different from those in other acts or WHoW dialogue acts? \end{itemize}

Finally, for each finalized sub-topic cluster, we manually assigned a label (e.g., “Echoing”) and a definition, iteratively refining them with GPT-4o using the cluster samples to accurately capture the content. When the output topics largely aligned with the original WHoW motive and dialogue act intersected labels, those original labels were retained. Additionally, we incorporated two intersected labels that appeared in more than 2.5\% but less than 10\% of instances, ensuring their relevance was not overlooked. Ultimately, this process yielded 10 classes of moderation strategies: four classes derived from the original WHoW intersected labels and six newly defined, domain-specific strategies.
 
\onecolumn
\section{Conversation club topics and material}
\label{app:soulreme}

% \usepackage{graphicx}
\begin{table}[H]
\resizebox{\columnwidth}{!}{%
\begin{tabular}{|p{0.2\linewidth}|p{0.80\linewidth}|}
\hline
Components     & Content                                                                                                            \\ \hline
Topic question & \begin{tabular}[c]{p{0.90\linewidth}}How AI impacts our daily life now and the future? \\ \begin{CJK}{UTF8}{gbsn}人工智能将会如何影响我们现在与未来的生活？\end{CJK}\end{tabular} \\ \hline
Description &
  \begin{tabular}[c]{p{0.9\linewidth}}In this discussion, we will explore the profound impacts of artificial intelligence on our daily lives, examining both current applications and future possibilities. As AI technologies advance, they integrate more seamlessly into various sectors such as healthcare, finance, education, and personal productivity, altering how we work, learn, and interact. This session aims to dissect the benefits and challenges of AI integration, and predict how it might shape our society in the coming decades.\\ \begin{CJK}{UTF8}{gbsn}在这次讨论中，我们将探讨人工智能对我们日常生活的深远影响，审视其当前的应用与未来的可能性。随着 AI 技术的进步，它正日益无缝融入医疗、金融、教育和个人生产力等各个领域，改变我们的工作、学习和互动方式。本次会议旨在剖析 AI 带来的机遇与挑战，并预测其在未来几十年内将如何塑造我们的社会和个人生活。\end{CJK}\end{tabular} \\ \hline
Questions &
  \begin{tabular}[c]{p{0.9\linewidth}}1. What are some of the most significant changes AI has brought to our personal and professional lives today? How do these changes enhance or complicate our daily activities?\\ \begin{CJK}{UTF8}{gbsn}人工智慧今天为我们的个人和职业生活带来了哪些重大变化？这些变化是如何影响我们的日常生活？\end{CJK}\\ \\ 2. AI is poised to automate many jobs that currently require human labor. Do you feel being threaten? What strategies should us individuals and the society implements to adapt?\\ \begin{CJK}{UTF8}{gbsn}人工智慧即将自动化许多目前需要人力的工作。你感到受到威胁了吗？我们个人和社会应该实施哪些策略来适应？\end{CJK}\\  \\ 3. How do you think AI might affect human relationships and social interactions in the future? Will it bring people closer or create more distance?\\ \begin{CJK}{UTF8}{gbsn}将来人工智慧会如何影响人际关係和社交互动？它会让人们更亲近还是造成更多距离？\end{CJK}\\  \\ 4. In comparison to AI, what are the things human possess that are irreplaceable?\\ \begin{CJK}{UTF8}{gbsn}与人工智慧相比，人类拥有哪些不可替代的特质？\end{CJK}\end{tabular} \\ \hline
\end{tabular}%
}
\caption{An exemplar discussion material featuring the topic “How AI impacts our daily life now and the future?” ncluding the main topic question, background description, and discussion prompts.}
\label{tab:session_material}
\end{table}

\begin{table}[H]
\resizebox{\columnwidth}{!}{%
\begin{tabular}{|p{0.2\linewidth}|p{0.8\linewidth}|}
\hline
Session stage            & Instruction                                                                                                                                 \\ \hline
Introduction(3 ms) & During the discussion stage, the moderator will briefly introduce the topic using the provided material and give a short self-introduction. \\ \hline
Discussion(~45 ms) &
  During the discussion stage, the moderator will guide the conversation using the provided questions, encouraging participants to share their thoughts and insights. If the discussion slows down, the moderator may contribute or introduce new prompts to stimulate engagement. \\ \hline
Conclusion(3 ms) &
  At the conclusion stage, the moderator will summarize the key points raised by participants for each question and provide a final thought. Additionally, the moderator will invite participants to share any final remarks. Finally, the session will be wrapped up with a closing statement and a farewell. \\ \hline
\end{tabular}%
}
\caption{The instruction for the moderator at different stages of the discussion session.}
\label{tab:session_instruction}
\end{table}

\section{Participation consent form}
\label{app:consent_form}
\makebox[\textwidth][c]{%
  \includegraphics[page=1,width=1.1\textwidth]{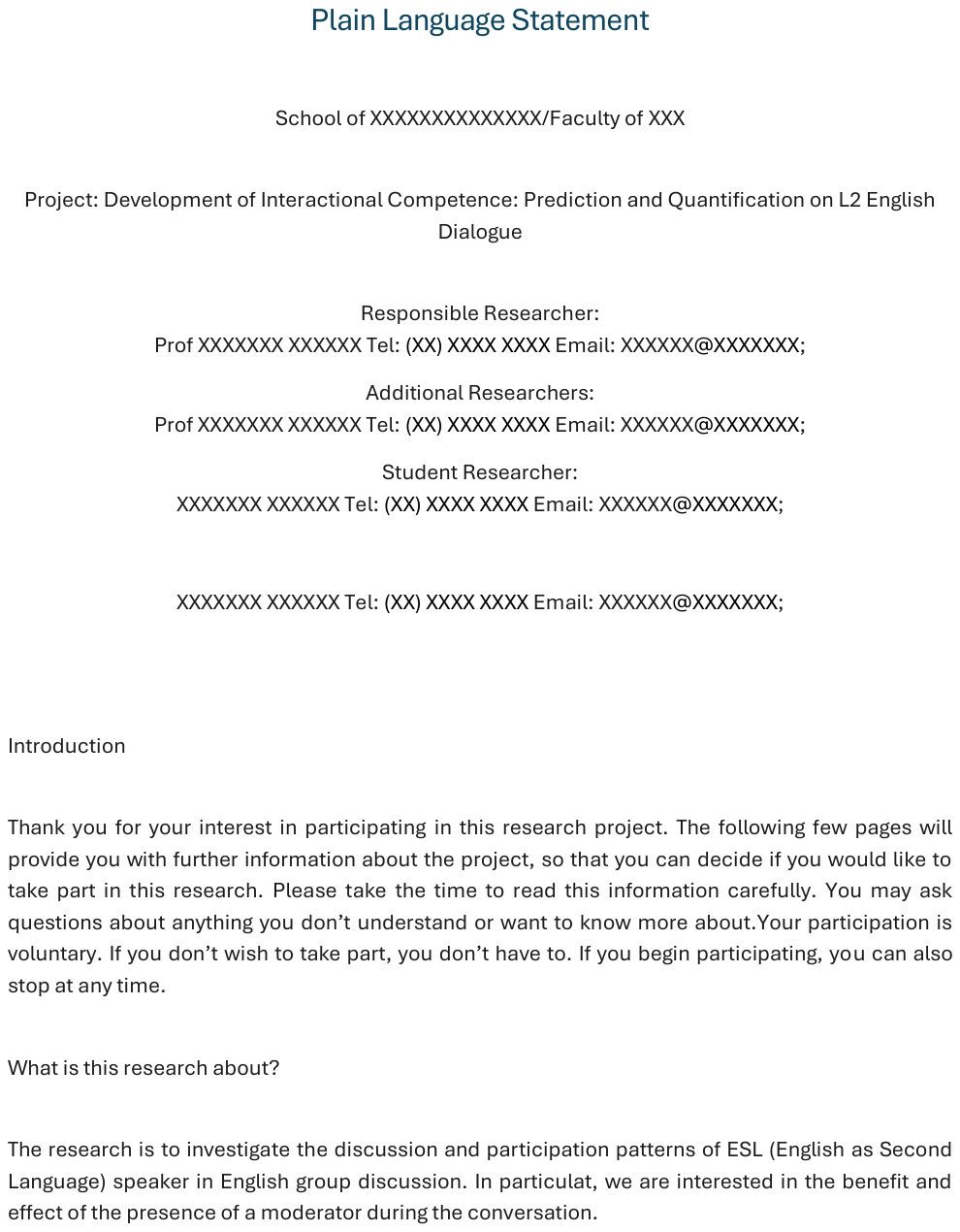}%
}
\newpage 
\makebox[\textwidth][c]{%
  \includegraphics[page=2,width=1.1\textwidth]{appendix/PLS_anonymised.pdf}%
}
\newpage 
\makebox[\textwidth][c]{%
  \includegraphics[page=3,width=1.1\textwidth]{appendix/PLS_anonymised.pdf}%
}
\newpage 
\makebox[\textwidth][c]{%
  \includegraphics[page=4,width=1.1\textwidth]{appendix/PLS_anonymised.pdf}%
}

\end{document}